# Real-Time Scheduling via Reinforcement Learning

**Robert Glaubius, Terry Tidwell, Christopher Gill, and William D. Smart**
Department of Computer Science and Engineering
Washington University in St. Louis
{rlg1@cse,ttidwell@cse,cdgill@cse,wds@cse}.wustl.edu

## Abstract

Cyber-physical systems, such as mobile robots, must respond adaptively to dynamic operating conditions. Effective operation of these systems requires that sensing and actuation tasks are performed in a timely manner. Additionally, execution of mission specific tasks such as imaging a room must be balanced against the need to perform more general tasks such as obstacle avoidance. This problem has been addressed by maintaining relative utilization of shared resources among tasks near a user-specified target level. Producing *optimal* scheduling strategies requires complete prior knowledge of task behavior, which is unlikely to be available in practice. Instead, *suitable* scheduling strategies must be learned online through interaction with the system. We consider the sample complexity of reinforcement learning in this domain, and demonstrate that while the problem state space is countably infinite, we may leverage the problem's structure to guarantee efficient learning.

## 1 Introduction

In cyber-physical systems such as mobile robots, setting and enforcing a utilization target for shared resources is a useful mechanism for striking a balance between general and mission-specific goals while ensuring timely execution of tasks. However, classical scheduling approaches are inapplicable to tasks in the domains we consider. First, some tasks are not efficiently preemptable: for example, actuation tasks involve moving a physical resource, such as a robotic arm or pan-tilt unit. Restoring the actuator state after a preemption would be essentially the same as restarting that task. Therefore, once an instance of a task acquires the resource, it should retain the resource until completion.

Second, the duration for which a task holds the resource may be stochastic. This is true for actuation tasks, which often involve one or more variable mechanical processes. Classical real-time scheduling approaches model tasks deterministically by treating a task's worst-case execution time (WCET) as its execution budget. This is inappropriate in our domain, as a task's WCET may be many orders of magnitude larger than its typical duration. To account for this variability, we assume that each task's duration obeys some underlying but unknown stationary distribution. Behaving optimally under these conditions requires that we account for this uncertainty in order to anticipate common events while exploiting early resource availability and hedging against delays.

In previous work (Glaubius et al., 2008, 2009), we have proposed methods for solving scheduling problems with these concerns, provided that accurate task models are available. One straightforward approach for employing these methods is via certainty equivalence: constructing and solving an approximate model from observations of the system. However, this is less effective than interleaving modeling and solution with execution, since interleaving learning allows the controller to adapt to conditions observed during execution, which may differ from conditions observed in a distinct modeling phase. Interleaving modeling and execution raises the *exploration/exploitation dilemma* (Kaelbling et al., 1996): the controller must balance optimal behavior with respect to available information against the long-term benefit of choosing apparently suboptimal exploratory actions that will improve that information. This dilemma is particularly relevant in the real-time systems domain, as sustained suboptimal behavior translates directly into poor quality of service.

In this paper we consider the problem of learning near-optimal schedules when the system model is not known in advance. We provide PAC bounds on the computational complexity of learning a near-optimal policy using balanced wandering. Our result is novel, as it extends established methods for learning in finite Markov decision processes to a domain with a countably infinite state space with unbounded costs. We also provide an empirical comparison of several exploration methods, and observe that the structure of the task scheduling problem enforces effective exploration.

## 2 Background

### 2.1 System Model

As in Glaubius et al. (2008, 2009), the task scheduling model consists of $n$ tasks $(T_i)_{i=1}^n$ that require mutually exclusive use of a single common resource. Each task $T_i$ consists of an infinite sequence of jobs $(T_{i,j})_{j=0}^\infty$. Job $T_{i,0}$ is available at time 0, while each job $T_{i,(j+1)}$ becomes available immediately upon completion of job $T_{i,j}$. Jobs cannot be preempted, so whenever a job is granted the resource, it occupies that resource for some stochastic duration until completion. Two simplifying assumptions are made regarding the distribution of job durations:

**(A1)** Inter-task job durations are independently distributed.

**(A2)** Intra-task job durations are independently and identically distributed.

When A1 holds, the duration of job $T_{i,j}$ always obeys the same distribution regardless of what job preceded it. This means that the system history is not necessary to predict the behavior of a particular job. When A2 holds, consecutive jobs of the same task obey the same distribution. Thus, every task $T_i$ has a *duration distribution* $P(\cdot|i)$ from which the duration of every job of $T_i$ is drawn. The actuator example in the previous section does not immediately satisfy these assumptions, since a job's duration depends on the state of the actuator when the job starts executing. These may be enforced in actuator-sharing, however, by requiring that each job leaves the actuator in a static reference position before relinquishing control.

In addition to the assumptions stated above, each duration distribution must have bounded support on the positive integers: that is, every task $T_i$ has an integer-valued WCET $W_i$ such that $\sum_{t=1}^{W_i} P(t|i) = 1$. For simplicity, $W$ denotes the maximum among all $W_i$, and the WCET of individual tasks are ignored.

Our goal is to schedule jobs in order to preserve *temporal isolation* (Srinivasan and Anderson, 2005) among tasks. We specify some target utilization $u_i$ for each task that describes its intended resource share at any temporal resolution. More specifically, let $x_i(t)$ denote the number of quanta during which task $T_i$ held the resource in the interval $[0, t)$. Our objective is to keep

$$|(t' - t)u_i - (x_i(t') - x_i(t))|$$

as small as possible over *every* time interval $[t, t')$ for *each* task $T_i$. We require that each task's utilization target $u_i$ is rational and that the resource is completely divided among all tasks, so that $\sum_{i=1}^n u_i = 1$.

### 2.2 MDP Formulation

Following Glaubius et al. (2008, 2009), this problem is modeled as a Markov decision process (MDP) (Puterman, 1994). An MDP consists of a set of states $\mathcal{X}$, a set of actions $\mathcal{A}$, a transition system $P$, and a cost function $C$. At each discrete decision epoch $k$, a controller observes the current MDP state $x_k$ and selects an action $i_k$. The MDP then transitions to state $x_{k+1}$ distributed according to $P(\cdot|x_k, i_k)$ and incurs cost $c_k = C(x_{k+1})$.

The value $V^\pi$ of a policy $\pi$ is the expected long-term $\gamma$-discounted cost of following $\pi$, where $\gamma$ is a discount factor in $(0, 1)$. $V^\pi$ satisfies the recurrence

$$V^\pi(x) = \sum_{y \in \mathcal{X}} P(y|x, \pi(x))[\gamma V^\pi(y) - C(y)].$$

It is often convenient to compare alternative actions using the state-action value function $Q^\pi(x, i)$,

$$Q^\pi(x, i) = \sum_{y \in \mathcal{X}} P(y|x, i)[\gamma V^\pi(y) - C(y)].$$

The objective is to find an optimal policy $\pi^*$ such that $V^{\pi^*}(x) \geq V^\pi(x)$ among all states $x$ and policies $\pi$. For brevity, $V$ and $Q$ are used to denote $V^{\pi^*}$ and $Q^{\pi^*}$. $V$ satisfies the *Bellman Equation* (Puterman, 1994)

$$V(x) = \max_{i \in \mathcal{A}} \sum_{y \in \mathcal{X}} P(y|x, i)[\gamma V(y) - C(y)], \quad (1)$$

or equivalently $V(x) = \max_i \{Q(x, i)\}$. An optimal policy is obtained by behaving greedily with respect to $Q$,

$$\pi^*(x) \in \mathrm{argmax}_i \{Q(x, i)\}.$$

Thus, computing the optimal control can be reduced to computing the optimal value function. Several dynamic programming and linear programming approaches have been developed to solve such problems when $\mathcal{X}$ and $\mathcal{A}$ are finite (Puterman, 1994).

The task scheduling problem is modeled as an MDP over a set of *utilization states* $\mathcal{X} = \mathbb{N}^n$. Each state $\mathbf{x}$ is an $n$-vector $\mathbf{x} = (x_1, \ldots, x_n)$ where each component $x_i$ is the total number of quanta during which task $T_i$ occupied the shared resource since system initialization. $\tau(\mathbf{x})$ denotes the total elapsed time in state $\mathbf{x}$,

$$\tau(\mathbf{x}) = \sum_{i=1}^{n} x_i. \qquad (2)$$

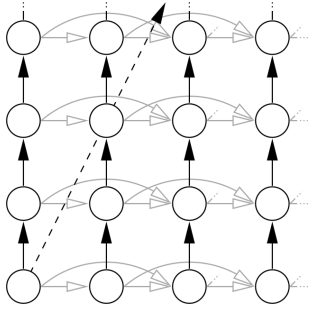

Figure 1: The utilization state model for a two-task problem instance. $T_1$ (grey, open arrowheads) stochastically transitions to the right, while $T_2$ (black, closed arrowheads) deterministically transitions upward. The dashed ray indicates the utilization target.

Each action $i$ in this MDP corresponds to the decision to run task $T_i$. Transitions are determined according to task duration distributions, so that

$$P(\mathbf{y}|\mathbf{x}, i) = \begin{cases} P(t|i) & \mathbf{y} = \mathbf{x} + t\Delta_i \\ 0 & otherwise \end{cases} \qquad (3)$$

where $\Delta_i$ is the zero vector except that component $i$ is equal to one, i.e., executing task $T_i$ alters just one dimension of the system state. The cost of a state is its $L_1$-distance from target utilization within the hyperplane of states with equal elapsed time $\tau(\mathbf{x})$,

$$C(\mathbf{x}) = \sum_{i=1}^{n} |x_i - \tau(\mathbf{x})u_i|. \qquad (4)$$

Figure 1 illustrates the utilization state model for a problem with two tasks and a target utilization $\mathbf{u} = (1, 2)/3$ (that is, task $T_1$ should receive 1/3 of the processor, and task $T_2$ should receive the rest). The target utilization defines a *target utilization ray* $\{\lambda\mathbf{u} : \lambda \geq 0\}$. When the components of $\mathbf{u}$ are rational, this ray regularly passes through many utilization states. In Figure 1, for example, the utilization ray passes through integer multiples of $(1, 2)$. Every state on this ray has zero cost, and states with the same displacement from the target utilization ray have equal cost.

This task scheduling MDP has an infinite state space and unbounded costs, but because of repeated transition and cost structure, states that are collinear along rays parallel to the utilization ray may be aggregated. The resulting problem still has infinitely many states, but an optimal policy can be estimated accurately using a finite state approximation (Glaubius et al., 2008). Applying this model minimization approach (Givan et al., 2003) does require prior knowledge of the task parameters, which is often unavailable in practice.

In this paper, we use reinforcement learning to integrate model and policy estimation. An important question is how much experience is necessary before we can trust learned policies. We address this question by deriving a PAC bound on the sample complexity of obtaining a near-optimal policy. To the best of our knowledge, this is the first such guarantee for problems with infinite state spaces and unbounded costs.

### 2.3 Related Work

A principle that unifies many successful methods for efficient exploration is *optimism in the face of uncertainty* (Kaelbling et al., 1996; Szita and Lőrincz, 2008). When presented with a choice between two actions with similar estimated value, methods using this principle tend to select the action that has been tried less frequently. Optimism can take the form of optimistic initialization (Even-Dar and Mansour, 2001), i.e., bootstrapping initial approximations of the value function with large values (Brafman and Tennenholtz, 2003; Strehl and Littman, 2008). *Interval estimation* techniques instead bias action selection towards exploration by maintaining confidence intervals on model parameters (Strehl and Littman, 2008; Auer et al., 2009) or value estimates (Even-Dar et al., 2002). Interval estimation techniques have been developed for solving single-state Bandit problems (Auer et al., 2002; Even-Dar et al., 2002; Mannor and Tsitsiklis, 2004; Mnih et al., 2008), as they can be extended to the general MDP setting by treating each state as a distinct Bandit problem.

Heuristic exploration strategies are often employed due to their relative simplicity. $\epsilon$-greedy exploration and Boltzmann action selection methods (Kaelbling et al., 1996) are randomization strategies that bias action selection toward exploitation. Perhaps the most commonly used strategy, $\epsilon$-greedy exploration, simply chooses an action uniformly at random with probability $\epsilon_k$ at epoch $k$, and otherwise it selects the apparent best action. By decaying $\epsilon_k$ appropriately this strategy asymptotically approaches the optimal policy (Even-Dar et al., 2002).

We are interested in quantifying the *sample complexity* of learning good policies in terms of the number of observations necessary to compute a near-optimal policy with high probability – i.e., probably approximately correct (PAC) learning (Valiant, 1984). Kakade (2003) has considered the question of PAC learning in MDPs in detail. Several PAC reinforcement learning algorithms have been developed, including $E^3$ (Kearns and Singh, 2002), R-Max (Brafman and Tennenholtz, 2003), MBIE (Strehl and Littman, 2008), and OIM (Szita and Lőrincz, 2008). These algorithms are limited to the finite state case, and assume bounded rewards. Metric $E^3$ (Kakade et al., 2003) is a PAC learner for MDPs with continuous but compact state spaces.

## 3 Online Learning Results

We consider the difficulty of learning good scheduling policies in this section. We approach this question both analytically and empirically. In Section 3.1, we derive a PAC bound (Valiant, 1984) on a balanced wandering approach to exploration (Kearns and Singh, 2002; Even-Dar et al., 2002; Brafman and Tennenholtz, 2003) in the scheduling domain. Our result is novel, as it extends results derived for the finite-state bounded cost setting, to a domain with a countably infinite state space and unbounded costs. These results rely on a specific Lipschitz-like condition that restricts the growth rate of the value function under our cost function (See Lemmas 3 and 4 in the appendix), and finite support of the duration distributions, *i.e.*, finite worst-case execution times of tasks. In Section 3.2, we present results from simulations comparing alternative exploration strategies.

We estimate task duration distributions using the empirical probability measure. We suppose a collection of $m$ observations $\{(i_k, t_k) : k = 1, \ldots, m\}$, where task $T_{i_k}$ ran for $t_k \sim P(\cdot|i_k)$ quanta at decision epoch $k$. Then let $\omega_m(i)$ be the number of observations involving task $T_i$, and let $\omega_m(i,t)$ be the number of those observations in which $T_i$ ran for $t$ quanta,

$$\omega_m(i) = \sum_{k=1}^{m} \mathbf{I}\{i_k = i\}, \tag{5}$$

$$\omega_m(i,t) = \sum_{k=1}^{m} \mathbf{I}\{i_k = i \wedge t_k = t\}, \tag{6}$$

where $\mathbf{I}\{\cdot\}$ is the indicator function. Then our task duration model $P_m(t|i)$ is just

$$P_m(t|i) = \omega_m(i,t)/\omega_m(i). \tag{7}$$

Since cost is completely determined by the system state, the transition model is the sole source of uncertainty in this problem.

### 3.1 Analytical PAC Bound

We consider the sample complexity of estimating a near-optimal policy with high confidence by bounding the number of low value exploratory actions taken (Kakade, 2003). Our analysis proceeds in three parts. First, we derive bounds on the value estimation error as a function of the model accuracy. Next, we determine the number of observations needed to guarantee that model accuracy. Finally, we use these results to determine how many observations suffice to arrive at a near-optimal policy with high certainty. We focus on estimating the state-action value function $Q$,

$$Q(\mathbf{x}, i) = \sum_{t=1}^{W} P(t|i)[\gamma V(\mathbf{x} + t\Delta_i) - C(\mathbf{x} + t\Delta_i)]. \tag{8}$$

We use $V_m$ to denote the optimal state value function and $Q_m$ to denote the state-action value function of the estimated MDP with transition dynamics $P_m$.

To establish our main result constraining the sample complexity of learning in our scheduling domain, we first provide the following simulation lemma, which is proven in the appendix.

**Lemma 1.** *If there is a constant $\beta$ such that for all tasks $T_i$,*

$$\sum_{t=1}^{W} |P_m(t|i) - P(t|i)| \leq \beta, \tag{9}$$

*where the worst-case execution time $W$ is finite, then*

$$\|Q_m - Q\|_\infty \leq \frac{2W\beta}{(1-\gamma)^2}. \tag{10}$$

This result serves an identical role to the Simulation Lemma of Kearns and Singh (2002) relating model estimation error to value estimation error. Our bound replaces the quadratic dependence on the number of states in that result with a dependence on the WCET $W$. This is consistent with observations indicating that the sample complexity of obtaining a good approximation should depend polynomially on the number of parameters of the transition model (Kakade, 2003; Leffler et al., 2007), which is $O(|\mathcal{X}|^2|\mathcal{A}|)$ for general MDPs, but is $(W \cdot |\mathcal{A}|)$ in this scheduling domain.

Theorem 1 provides a PAC bound on the number of observations needed to arrive at an accurate estimate of the value function. For the sake of simplicity we assume balanced wandering here, as this result can be easily used to guide offline modeling as well as employed during online learning.

**Theorem 1.** *Under balanced exploration, if*

$$m \geq \left(\frac{32W^3 n}{\varepsilon^2(1-\gamma)^4}\right) \log\left(\frac{2Wn}{\delta}\right), \tag{11}$$

*then $\|Q_m - Q\|_\infty \leq \varepsilon$ with probability at least $1 - \delta$.*

*Proof.* According to Lemma 1, model accuracy $\beta \leq \varepsilon(1-\gamma)^2/(2W)$ is sufficient to guarantee that $\|Q_m - Q\|_\infty \leq \varepsilon$. Thus, demonstrating the bound in Equation 11 is a matter of guaranteeing with high certainty that $P_m$ is near $P$; specifically, we require that

$$\mathbf{P}\Big\{\bigwedge_{i=1}^{n}\Big(\sum_{t=1}^{W} |P_m(t|i) - P(t|i)| > \beta\Big)\Big\} \leq \delta,$$

which we can enforce using the union bound by requiring $\mathbf{P}\left\{\sum_{t=1}^{W} |P_m(t|i) - P(t|i)| \leq \beta\right\} \geq 1 - \delta/n$ for every task. By Lemma 8.5.5 from Kakade's dissertation (Kakade, 2003),

$$\omega_m(i) \geq (8W/\beta^2)\log(2Wn/\delta)$$

is sufficient to guarantee with probability $1 - \delta/n$ that $P_m(\cdot|i)$ is accurate. If we assume balanced wandering, that $\omega_m(i) = m/n$ for each task $T_i$, then we require

$$m \geq (8Wn/\beta^2)\log(2Wn/\delta) \qquad (12)$$

observations. Substituting the least accuracy $\beta = \varepsilon(1-\gamma)^2/(2W)$ that will still guarantee an $\varepsilon$-approximation to $Q$, produces the stated result,

$$m \geq \left(\frac{32W^3 n}{\varepsilon^2(1-\gamma)^4}\right)\log\left(\frac{2Wn}{\delta}\right). \qquad \square$$

Theorem 1 provides a PAC bound on the number of observations needed to learn an $\varepsilon$-approximation to $Q$. However, we are principally interested in discovering the number of observations we need to trust our learned policies. Corollary 1 establishes the sample complexity for using balanced complexity to learn good scheduling policies.

**Corollary 1.** *Assuming each action is tried an equal number of times, if*

$$m \geq \left(\frac{128W^3\gamma^2 n}{\varepsilon^2(1-\gamma)^6}\right)\log\left(\frac{2Wn}{\delta}\right),$$

*then the optimal policy $\pi_m$ of the estimated task scheduling MDP is within $\varepsilon$ of the optimal policy $\pi$ with probability at least $1 - \delta$.*

A classical result due to Singh and Yee (1994) demonstrates that, in general, a policy $\hat{\pi}$ that is greedy with respect to value function approximation $\hat{V}$ is within $2\gamma\|\hat{V} - V\|_\infty/(1-\gamma)$ of optimal. Corollary 1 follows by noting that $\|\hat{V} - V\|_\infty \leq \|\hat{Q} - Q\|_\infty$, so we require that $2\gamma\|\hat{Q} - Q\|_\infty/(1-\gamma) \leq \varepsilon$. Substituting this constraint on $Q_m$ into Theorem 1 establishes the corollary.

As with existing bounds, the sample complexity scales polynomially in the parameters $1/\gamma$, $1/\delta$, $1/\varepsilon$, and the number of actions. Unlike bounds for general MDPs, there is no dependence on the number of states; instead, the complexity of learning is determined by the worst-case execution time $W$. This result is similar to bounds for relocatable action models (Leffler et al., 2007), in which the state space can be partitioned into a relatively small number of classes. Transition models can be generalized among states in the same class, so the sample complexity of learning depends on the number of classes rather than the number of states. Our scheduling MDP is a special case of the relocatable action model in which there is only one class of states.

While relocatable action models have been used to address infinite state spaces (Brunskill et al., 2009), existing sample complexity results do not address the unbounded reward case. We are able to handle unbounded costs here by taking advantage of the slow growth rate of the value function relative to the discount factor. Specifically, the distance between consecutive states is bounded, so while costs grow polynomially with distance from the resource share target (cf. Lemma 3 in the appendix), since costs are exponentially discounted the value of any particular state is finite. These observations enable the bound in Lemma 1, suggesting that sample complexity bounds may be possible in general for infinite state, unbounded cost models as long as the number of classes is finite and individual state values can be bounded. Of course, for these results to be useful good policies must be represented compactly, which is possible for the scheduling domain considered here (Glaubius et al., 2008), but is not generally the case.

### 3.2 Empirical Evaluation

The PAC bound in the previous section gives a sense of the finite sample performance for learning a good policy; however, it requires several simplifying assumptions, such as balanced wandering, so the bound may not be tight. In practice, alternative exploration strategies may yield better performance than our bound would indicate. We compare the performance of several exploration strategies in the context of the task scheduling problem by conducting experiments comparing $\epsilon$-greedy, balanced wandering, and an interval-based exploration strategy.

For interval-based optimistic exploration, we use the confidence intervals derived for the multi-armed bandit case by Even-Dar et al. (2002) for the Successive Elimination algorithm. That algorithm constructs intervals of the form $\alpha_k = \sqrt{\log(nk^2 c)/k}$ about the expected cost of each action at decision epoch $k$, then eliminates actions that appear worse than the apparent best using an overlap test. The parameter $c$ controls the sensi-

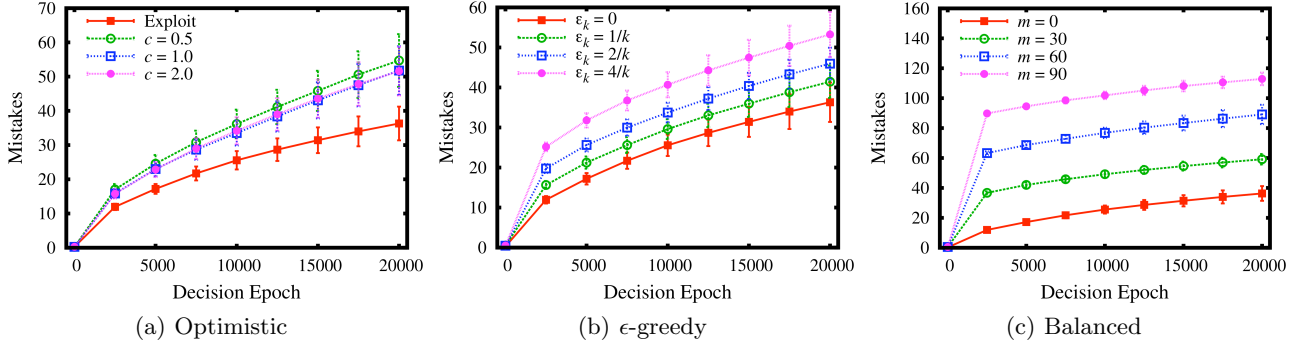

Figure 2: Simulation comparison of exploration techniques. Note the differing scales on the vertical axes.

tivity of the intervals. We use these intervals to select actions optimistically according to

$$\operatorname*{argmax}_{i \in \mathcal{A}} \{Q_m(\mathbf{x}, i) + \alpha_{i,k}\}$$

where we have adjusted the confidence intervals according to the potentially different number of observations of each task,

$$\alpha_{k,i} = \sqrt{\log(n\omega_k(i)^2 c)/\omega_k(i)}$$

We vary $c$ to control the chance of taking exploratory actions. As $c$ shrinks, these intervals narrow, increasing the tendency to exploit the estimated model.

In our experiments with $\epsilon$-greedy, we set the random selection rate at decision epoch $k$, $\epsilon_k = \epsilon_0/k$ for varying values of $\epsilon_0$; this strategy always exploits when $\epsilon_0 = 0$. Balanced wandering simply executes each task a fixed number of times $m$ prior to exploiting. We vary this parameter to determine its impact on the learning rate. When $m = 0$, this strategy always exploits its current model knowledge.

To compare the performance of these exploration strategies, we generated 400 random problem instances with two tasks. Duration distributions for these tasks were generated by first selecting a worst-case execution time $W$ uniformly at random from the interval $[8, 32]$, then choosing a normal distribution with mean and variance selected uniformly at random from the respective intervals $[1, W]$ and $[1, 4]$; this distribution was then truncated and discretized over the interval $[1, W]$. Utilization targets for each task were chosen according to $\mathbf{u} = (u'_1, u'_2)/(u'_1 + u'_2)$, where $u'_1$ and $u'_2$ were integers selected uniformly at random between $[1, 64]$. We used a discount factor of $\gamma = 0.95$ in our tests.

We conducted experiments by initializing the model in the state $\mathbf{x} = (0, 0)$. The controller simulated a single trajectory over 20,000 decision epochs in each problem instance with each exploration strategy. In order to avoid enumerating arbitrarily large numbers of states, we reinitialized the state whenever a state with cost greater than 50 was encountered. These high cost states were treated as absorbing states in the approximate model to avoid degenerate policies that exploit the reset. We report the number of mistakes – the number of times the exploration strategy chooses an suboptimal action $i$ that has value $V(\mathbf{x}) - Q(\mathbf{x}, i) \geq 10^{-6}$. The results of these experiments are shown in Figure 2.

### 3.3 Evaluation Results

In Figure 2, we report 90% confidence intervals on the mean number of mistakes each exploration strategy makes, averaged across the problem instances described above. Note that these plots have different scales due to the variation in mistake rates among exploration strategies.

Figure 2(a) compares the performance of interval-based optimistic action selection to that of "Exploit", the policy that greedily follows the optimal policy of the approximate model at each decision epoch. All of the interval-based exploration settings we considered exhibited statistically similar performance. Interestingly, the exploitive strategy yields better performance than the explorative strategies despite its lack of an explicit exploration mechanism.

This observation holds true for $\varepsilon$-greedy exploration and balanced wandering as well. Figure 2(b) illustrates the performance of $\epsilon$-greedy exploration. Notice that the mistake rate decreases along with the likelihood of taking exploratory actions – that is, as $\epsilon_0$ approaches zero. Explicit exploration may not improve performance in this domain. This is further supported by our results for balanced wandering. The theory behind balanced wandering is that making a few initial mistakes early on will pay off in the long run due to more uniformly accurate models. Figure 2(c) shows that this is not the case in our scheduling domain, as a purely exploitive strategy $m = 0$ outperforms each balanced wandering approach.

These results suggest that the exploitive strategy may be the best available exploration method in our task scheduling problem domain. One plausible explanation is that the environment itself enforces rational exploration: if some task is never dispatched, the system will enter progressively more costly states as that task becomes more and more underused. Thus, eventually the estimated benefit of running that task will be substantial enough that the exploitive strategy must use it. It is interesting to note that all of the explorative policies considered have quite low mistake rates despite the tight threshold of $10^{-6}$ used to distinguish suboptimal actions.

## 4 Conclusions

In this paper we have considered the problem of learning near-optimal schedules when the system model is not fully known in advance. We have presented analytical results that bound the number of suboptimal actions taken prior to arriving at a near-optimal policy with high certainty. Interestingly, the transition system's portability results in bounds that are similar to those for estimating the underlying model in a single state.

This naturally leads to a comparison to the multi-armed bandit model (see, for example, Even-Dar et al. (2002)), in which there is a single state with several available actions. Each action causes the emission of a reward according to a corresponding unknown, stationary random process. However, a bandit model does not appear to apply directly because while the duration distributions are stationary processes that are invariant between states, the payoff associated with each action is state-dependent.

We have focused on the PAC model of learning rather than deriving bounds on *regret* – the loss in value incurred due to suboptimal behavior during learning (Auer et al., 2009). Regret bounds may translate more readily into guarantees about transient real-time performance effects during learning, as guarantees regarding cost (and hence value) translate into guarantees about task timeliness.

We have presented empirical results which suggest that a learner that always exploits its current information outperforms agents that explicitly encourage exploration in this domain. This occurs because any policy that consistently ignores some action will get progressively farther from the utilization target, resulting in arbitrarily large costs. Thus the domain itself appears to enforce an appropriate level of exploration, perhaps obviating the need for an explicit exploration mechanism. It is an open question whether a more general class of MDPs that exhibit this behavior can be identified.

### Acknowledgements

This research has been supported in part by NSF grants CNS-0716764 (Cybertrust) and CCF-0448562 (CAREER).

## Appendix: Proof of Lemma 1

Lemma 1 states that the error in approximating $Q$ is bounded,

$$\|Q_m - Q\|_\infty \leq 2W\beta/(1-\gamma)^2,$$

when the transition model estimation error is bounded by $\beta$ (cf. Equation 9), where $W$ is the maximum worst-case execution time among all tasks. We introduce lemmas prior to demonstrating this result. The first provides a bound on expected successor state value of a function with a Lipschitz-like "speed limit" on its growth. Subsequent lemmas establish that both costs and values exhibit this property.

**Lemma 2.** *Suppose $p$ and $\hat{p}$ are distributions over $\{1, \ldots, W\}$ that satisfy $\sum_{t=1}^{W} |p(t) - \hat{p}(t)| \leq \beta$, and that for any $i$, the function $f : \mathcal{X} \to \mathbb{R}$ satisfies $|f(\mathbf{x}_{i,t}) - f(\mathbf{x})| \leq \lambda t$ for some $\lambda \geq 0$. Then*

$$\Big|\sum_{t=1}^{W}[p(t) - \hat{p}(t)]f(\mathbf{x} + t\Delta_i)\Big| \leq \lambda W \beta.$$

*Proof.* Since we can decompose $f(\mathbf{x} + t\Delta_i)$ into an $f(\mathbf{x})$ term and a $\lambda t$ term, we have

$$\Big|\sum_t [p(t) - \hat{p}(t)]f(\mathbf{x} + t\Delta_i)\Big|$$
$$\leq \Big|\sum_t [p(t) - \hat{p}(t)]f(\mathbf{x})\Big| + \lambda \sum_t |p(t) - \hat{p}(t)|\, t.$$

Since $f(\mathbf{x})$ does not depend on $t$, the first term on the right-hand side vanishes. Since $t \leq W$, we have

$$\lambda \sum_t |p(t) - \hat{q}(t)|\, t \leq \lambda W \beta. \qquad \square$$

We now show that the cost function $C$ and the optimal value $V$ satisfy the conditions of Lemma 2.

**Lemma 3.** *For any state $\mathbf{x}$, task $T_i$, and duration $t$,*

$$|C(\mathbf{x}) - C(\mathbf{x} + t\Delta_i)| \leq t C(\Delta_i). \qquad (13)$$

*Proof.* Since $C(\mathbf{x})$ is the $L_1$-norm between $\mathbf{x}$ and $\tau(\mathbf{x})\mathbf{u}$ (cf. Equation 4), we can use the triangle inequality and scalability to derive the upper bound

$$C(\mathbf{x} + t\Delta_i) \leq C(\mathbf{x}) + tC(\Delta_i).$$

We can also use the triangle inequality to obtain the lower bound, since

$$C(\mathbf{x}) \leq C(\mathbf{x} + t\Delta_i) + tC(\Delta_i);$$

rearranging the terms yields the intended result. □

It is straightforward to show that $C(\Delta_i) < 2$ for any task $T_i$. We make use of this fact and Lemma 3 to derive a related limit on the growth of $V$.

**Lemma 4.** *For any state $\mathbf{x}$, task $T_i$, and duration $t$,*

$$|V(\mathbf{x} + t\Delta_i) - V(\mathbf{x})| \leq 2t/(1-\gamma).$$

*Proof.* Let $\mathbf{y} = \mathbf{x} + t\Delta_i$. We can bound the difference in values at $\mathbf{x}$ and $\mathbf{y}$ in terms of the difference in $Q$-values, since

$$|V(\mathbf{y}) - V(\mathbf{x})| \leq \max_j |Q(\mathbf{y}, j) - Q(\mathbf{x}, j)|. \qquad (14)$$

By expanding $Q$ according to Equation 8 and rearranging terms,

$$\begin{aligned}
&|Q(\mathbf{y}, j) - Q(\mathbf{x}, j)| \\
&= \Big|\sum_s P(s|j)\big(\gamma V(\mathbf{y} + s\Delta_j) - \gamma V(\mathbf{x} + s\Delta_j) \\
&\qquad - C(\mathbf{y} + s\Delta_j) + C(\mathbf{x} + s\Delta_j)\big)\Big| \\
&\leq \gamma \sum_s P(s|j) |V(\mathbf{y} + s\Delta_j) - V(\mathbf{x} + s\Delta_j)| \\
&\qquad + \sum_s P(s|j) |C(\mathbf{y} + s\Delta_j) - C(\mathbf{x} + s\Delta_j)| \\
&\leq 2t + \gamma \sum_s P(s|j) |V(\mathbf{y} + s\Delta_j) - V(\mathbf{x} + s\Delta_j)|.
\end{aligned}$$

Recurring this argument on the absolute value in the right-hand side results in accumulating a residual $\gamma^k tC(\Delta_i)$ for the $k^{th}$ repetition. Therefore,

$$|V(\mathbf{x} + t\Delta_i) - V(\mathbf{x})| \leq \sum_{k=0}^{\infty} \gamma^k tC(\Delta_i) = \frac{2t}{1-\gamma}. \quad \square$$

We are ready now to prove Lemma 1.

*Proof of Lemma 1.* We begin bounding $|Q(\mathbf{x}, i) - Q_m(\mathbf{x}, i)|$ by expanding according to Equation 8, rearranging terms to group costs and values, then decomposing the sum by using the superadditivity of the absolute value:

$$\begin{aligned}
&|Q(\mathbf{x}, i) - Q_m(\mathbf{x}, i)| \\
&\leq \gamma \Big|\sum_t P(t|i)V(\mathbf{x} + t\Delta_i) - P_m(t|i)V_m(\mathbf{x} + t\Delta_i)\Big| \\
&\quad + \Big|\sum_t [P(t|i) - P_m(t|i)]C(\mathbf{x} + t\Delta_i)\Big|. \qquad (15)
\end{aligned}$$

Applying Lemmas 2 and 3, we have

$$\Big|\sum_t [P(t|i) - P_m(t|i)]C(\mathbf{x} + t\Delta_i)\Big| \leq 2W\beta.$$

We can apply the triangle inequality to obtain

$$\begin{aligned}
&\Big|\sum_t P(t|i)V(\mathbf{x} + t\Delta_i) - P_m(t|i)V_m(\mathbf{x} + t\Delta_i)\Big| \\
&\leq \Big|\sum_t [P(t|i) - P_m(t|i)]V(\mathbf{x} + t\Delta_i)\Big| \\
&\quad + \sum_t P_m(t|i) |V(\mathbf{x} + t\Delta_i) - V_m(\mathbf{x} + t\Delta_i)|.
\end{aligned}$$

Using Lemmas 2 and 4 yields

$$\Big|\sum_t [P(t|i) - P_m(t|i)]V(\mathbf{x} + t\Delta_i)\Big| \leq 2W\beta/(1-\gamma).$$

Substituting back into Equation 15 allows us to write

$$\begin{aligned}
|Q(\mathbf{x}, i) - Q_m(\mathbf{x}, i)| &\leq 2W\beta + \gamma \frac{2W\beta}{1-\gamma} \\
&\quad + \gamma \sum_t P_m(t|i) |V(\mathbf{x} + t\Delta_i) - V_m(\mathbf{x} + t\Delta_i)|.
\end{aligned}$$

Finally, we can use Equation 14 to express $|V(\mathbf{x} + t\Delta_i) - V_m(\mathbf{x} + t\Delta_i)|$ in terms of $Q$, then recur this argument to produce the stated bound,

$$|Q(\mathbf{x}, i) - Q_m(\mathbf{x}, i)| \leq \sum_{k=0}^{\infty} \gamma^k \frac{2W\beta}{1-\gamma} = \frac{2W\beta}{(1-\gamma)^2}. \quad \square$$